\RequirePackage{fix-cm}
\documentclass[10pt]{lynnreal}
\usepackage{lmodern}
\usepackage{adjustbox}
\microtypesetup{expansion=true}

\usepackage{amsmath,amssymb}
\usepackage{array}
\usepackage{booktabs}
\usepackage{multirow}
\usepackage{float}
\usepackage{subcaption}
\usepackage{cleveref}
\graphicspath{{figures/}}

\newcommand{\resulttablestyle}{%
  \scriptsize
  \setlength{\tabcolsep}{2pt}%
  \renewcommand{\arraystretch}{1.05}%
}

\title{\LARGE HandiVid: A High-Quality Dataset for Hand-Driven\\Interactive Video Generation}
\author[1,2,3,*]{Chaoyang Zhang}
\author[1,2,4,*,\dagger]{Shaohao Rui}
\author[1,2,5,*,\dagger]{Xiaofeng Mao}
\author[1,2,6]{Yuanxun Xu}
\author[2]{Haoyuan Xia}
\author[2,4]{Xiaohong Liu}
\author[3]{Qinghe Du}
\author[1,2,\ddagger]{Weijie Ma}

\affiliation[1]{LynnReal AI}
\affiliation[2]{Shanghai Innovation Institute}
\affiliation[3]{Xi'an Jiaotong University}
\affiliation[4]{Shanghai Jiaotong University}
\affiliation[5]{Fudan University}
\affiliation[6]{Harbin Institute of Technology}

\contribution[*]{Equal contribution}
\contribution[\dagger]{Project leader}
\contribution[\ddagger]{Corresponding author}
\lynndata[Code]{\url{https://github.com/LynnReal-AI/LynnReal}}
\lynndata[Database]{\url{https://huggingface.co/datasets/shaohao011/handivid}}

\abstract{Hand-driven egocentric interactive video generation aims to generate dynamic scenes with realistic interaction responses conditioned on hand movements and camera motion. However, existing egocentric video datasets struggle to provide scene diversity, meaningful hand-object interactions, and reliable motion annotations simultaneously. To address these limitations, we introduce \textbf{HandiVid}, a high-quality dataset comprising 200K curated and annotated egocentric interaction video clips for hand-driven interactive video generation. We develop a unified data processing pipeline that integrates public datasets, online videos and synthetic videos. Through video quality assessment, hand detection filtering, and multi-stage interaction validation using vision-language models (VLMs), we select high-quality video clips with meaningful hand-object interactions. Since 3D hand reconstruction relies on initial hand detections, we integrate SAM 3 into the hand detection pipeline to reduce missed detections and improve the completeness of 3D hand annotations. Each video clip is provided with high-quality hand renderings, camera trajectories, and interaction descriptions. Comprehensive dataset analysis and downstream video generation experiments demonstrate the effectiveness of \textbf{HandiVid} in terms of data diversity, annotation quality, and controllable video generation.}

\begin{document}
\maketitle

\begin{figure}[t]
  \centering
  \includegraphics[width=\linewidth]{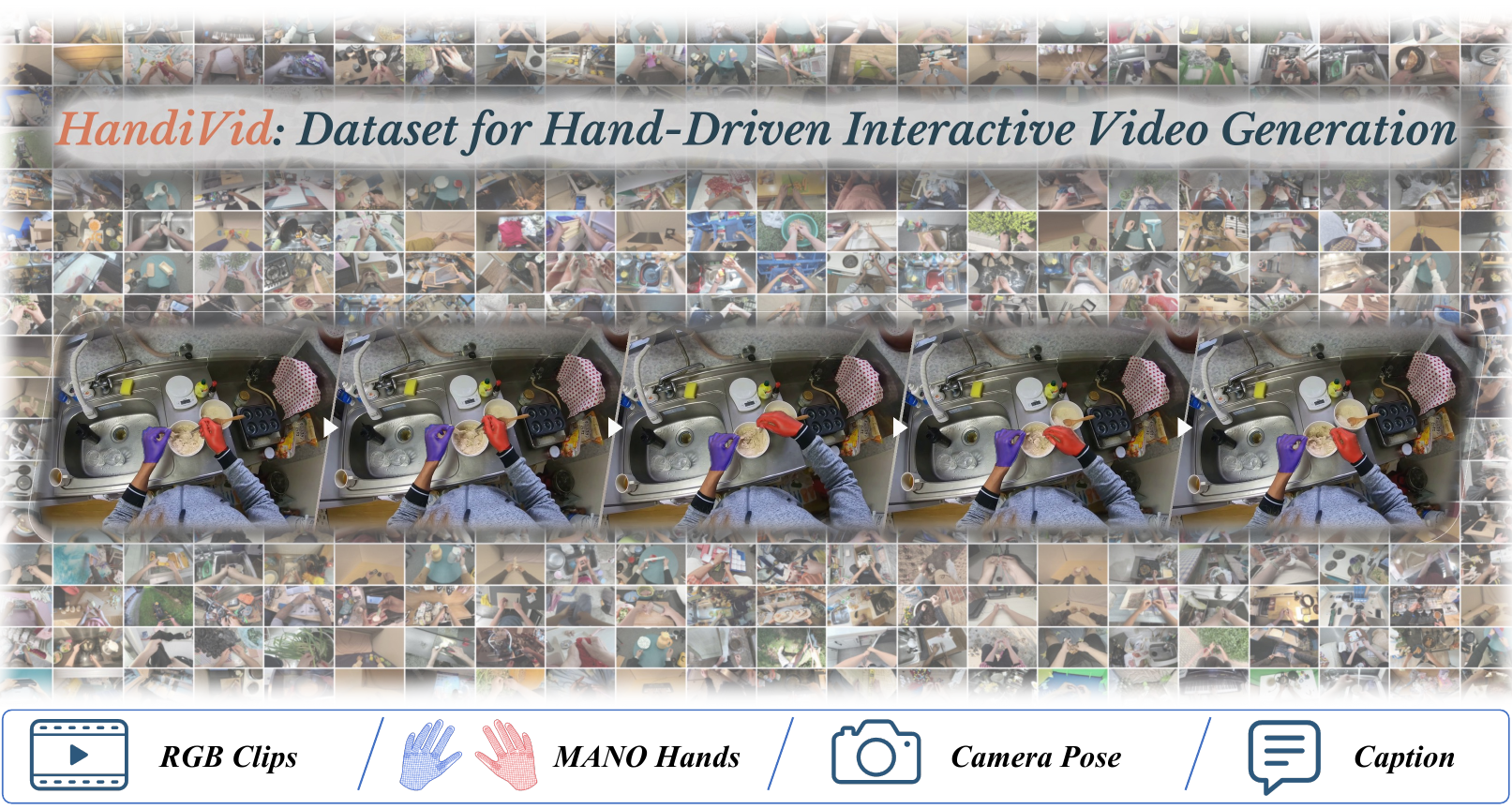}
  \caption{\textbf{HandiVid} provides 200K curated egocentric interaction clips with synchronized 3D hand renderings, camera trajectories, and interaction descriptions, which support training hand-controlled interactive video generation models. Videos from public datasets, online sources, and synthetic videos are processed through a unified filtering and annotation pipeline.}
  \label{fig:teaser}
\end{figure}

\section{Introduction}
\label{sec:introduction}

Hand-driven egocentric video generation aims to synthesize scene responses to hand actions and viewpoint changes, enabling interactive virtual environments. Recent methods such as Hand2World~\cite{wang2026hand2world} and Generated Reality~\cite{xie2026generatedreality} jointly use hand and camera motion to control generation. The generated sequence must preserve local manipulation details, including finger contact and object motion, while maintaining scene consistency as the viewpoint changes. Learning these responses requires videos that clearly capture hand-object manipulation across diverse scenes, paired with synchronized motion annotations and interaction descriptions.

Existing datasets provide complementary but incomplete supervision. Ego4D~\cite{grauman2022ego4d} and EPIC-KITCHENS-100~\cite{damen2022rescaling} capture diverse activities but lack consistent 3D hand and camera annotations. ARCTIC~\cite{fan2023arctic}, H2O~\cite{kwon2021h2o}, and TACO~\cite{liu2024taco} provide detailed geometric annotations but are limited in capture settings and object diversity. EgoDex~\cite{hoque2026egodex} and Open-AoE~\cite{li2026openaoe} offer larger-scale manipulation data with language and motion annotations. Their episode and action boundaries may not match the temporal extent of short generation clips, so descriptions can refer to actions outside the visible interval. Integrating these datasets thus requires quality filtering, interaction verification, motion representation unification, and clip-aligned descriptions.

Generation-oriented datasets partially address these limitations. EgoVid-5M~\cite{wang2024egovid} provides curated egocentric videos with textual descriptions and estimated camera motion, but lacks native 3D hand annotations. HandsOnWorld~\cite{chen2026handsonworld} introduces EgoVid-Pro with hand trajectories obtained through detection and geometric filtering. It selects interactions through action word matching, while geometric checks remove other people's hands and unstable trajectories. However, hand visibility and action labels do not necessarily indicate meaningful object interactions, as clips may contain hovering hands or static objects without actual manipulation. Effective training data should capture purposeful hand actions, such as object placement or tool use, along with the resulting object motion or state changes. Explicit interaction verification is therefore essential to identify clips that capture the relationship between hand actions and their effects on the environment.

We introduce \textbf{HandiVid}, a dataset of approximately 200K egocentric interaction clips collected from public datasets, online videos, and model-generated content (\cref{fig:teaser}). It covers diverse scenarios, including everyday object manipulation, tool use, and bimanual coordination, across varied environments and recording conditions. Each clip is annotated with 3D hand parameters, frame-aligned hand renderings, camera trajectories, and clip-level interaction descriptions. To ensure annotation quality and consistency, we develop a unified processing pipeline that integrates visual quality filtering, hand motion reconstruction, camera trajectory estimation, and captioning. Specifically, we use SAM 3~\cite{carion2025sam3} to recover missing hand regions and improve reconstruction completeness, and employ vision-language models (VLMs) to verify meaningful hand-object interactions. We generate descriptions from the cropped clips so that language and motion annotations cover the same temporal interval. We further filter unreliable camera trajectories and balance scene and activity distributions through targeted sampling. Together, these steps yield a diverse, interaction-focused dataset with consistent motion annotations and descriptions.

Our contributions are as follows:
\begin{itemize}
    \item We introduce \textbf{HandiVid}, a diverse egocentric interaction dataset with aligned hand, camera, and language conditions for controllable video generation.
    \item We develop a unified filtering and annotation pipeline with hand detection completion, VLM-based interaction verification, and sampling by scene and activity.
    \item We assess annotation quality and demonstrate improved video generation quality and motion consistency over a baseline trained on ARCTIC, including out-of-distribution generalization evaluation.
\end{itemize}

\section{Related work}
\label{sec:related}

\subsection{World generation models}

Recent video generation models~\cite{yang2025cogvideox,kong2024hunyuanvideo,wan2025wan,minimax2026h3} have advanced visual synthesis, providing a foundation for world modeling. Genie~\cite{bruce2024genie} and Cosmos~\cite{nvidia2025cosmos} explore controllable world simulation, while CausVid~\cite{yin2025causvid}, Self Forcing~\cite{huang2025selfforcing}, and SolarWM~\cite{huang2026solarwm} investigate autoregressive generation and long-term temporal consistency~\cite{mao2026lynnrealomni,rui2026biwm,mao2026packforcing}.

Motion conditioning further enables controllable video generation. MotionCtrl~\cite{wang2024motionctrl}, CameraCtrl~\cite{he2024cameractrl}, and ReCamMaster~\cite{bai2025recammaster} focus on camera and object motion control, while PlayerOne~\cite{tu2025playerone} and PEVA~\cite{bai2025peva} incorporate human actions into egocentric generation. For hand-object interactions, CosHand~\cite{sudhakar2024coshand}, InterDyn~\cite{akkerman2025interdyn}, and Mask2IV~\cite{li2026mask2iv} explore action-conditioned prediction and interaction synthesis. Re-HOLD~\cite{fan2025rehold} and SpriteHand~\cite{li2025spritehand} further investigate interaction reenactment and real-time generation.

Recent methods introduce 3D hand and camera conditioning for more precise control. Dexterous World Models~\cite{kim2025dwm} and occlusion-aware 3D joint conditioning~\cite{zhang2026occlusion} leverage hand motion to guide generation, while Hand2World~\cite{wang2026hand2world} and Generated Reality~\cite{xie2026generatedreality} jointly model hand actions and viewpoint changes. EgoHOI~\cite{li2026egohoi}, HandWorld~\cite{sun2026handworld}, HandsOnWorld~\cite{chen2026handsonworld}, and EgoSim~\cite{hao2026egosim} further explore hand-object interactions and scene dynamics. These approaches require interaction videos paired with reliable motion annotations. To support this need, HandiVid provides diverse interaction clips with aligned hand motion, camera trajectories, and interaction descriptions.

\subsection{Datasets for hand-driven video generation}

Egocentric activity datasets capture natural interactions with semantic annotations. Ego4D~\cite{grauman2022ego4d} covers diverse daily activities, EPIC-KITCHENS-100~\cite{damen2022rescaling} provides fine-grained action narrations of kitchen activities, and CaptainCook4D~\cite{peddi2024captaincook4d} annotates cooking procedures and execution errors. However, these datasets primarily focus on activity understanding and procedural analysis, often containing clips without meaningful manipulation and lacking consistent hand and camera motion annotations required for controllable video generation.

Hand-object interaction datasets provide rich geometric supervision. H2O~\cite{kwon2021h2o} and HOI4D~\cite{liu2022hoi4d} offer hand-object poses and action annotations, while ARCTIC~\cite{fan2023arctic} captures bimanual manipulation and contact. HOT3D~\cite{banerjee2025hot3d} supports hand-object tracking, while TACO~\cite{liu2024taco} covers diverse tool interactions. However, their capture settings and object sets remain limited, motivating the integration of complementary sources for broader interaction coverage.

Large-scale manipulation datasets provide motion and language supervision for embodied learning. EgoDex~\cite{hoque2026egodex} collects 829 hours of Apple Vision Pro recordings across 194 tabletop tasks, with hand joints, camera poses, and task language. Open-AoE~\cite{li2026openaoe} records natural manipulation using consumer smartphones and provides MANO hand poses, camera trajectories, and temporally bounded atomic action descriptions for action-conditioned video and world model training.

For egocentric video generation, EgoVid-5M~\cite{wang2024egovid} builds on Ego4D to provide short clips, automatic descriptions, and motion information. Its filtering uses text-video alignment, motion, and quality scores. HandsOnWorld~\cite{chen2026handsonworld} constructs EgoVid-Pro by recovering the camera wearer's hand trajectories through detection and 3D geometric filtering. HandiVid directly verifies purposeful manipulation from video content and combines visual quality filtering, unified motion annotation, and sampling by scene and activity to curate videos from multiple sources.

\section{Dataset construction}
\label{sec:construction}

\begin{figure}[t]
  \centering
  \includegraphics[width=\textwidth]{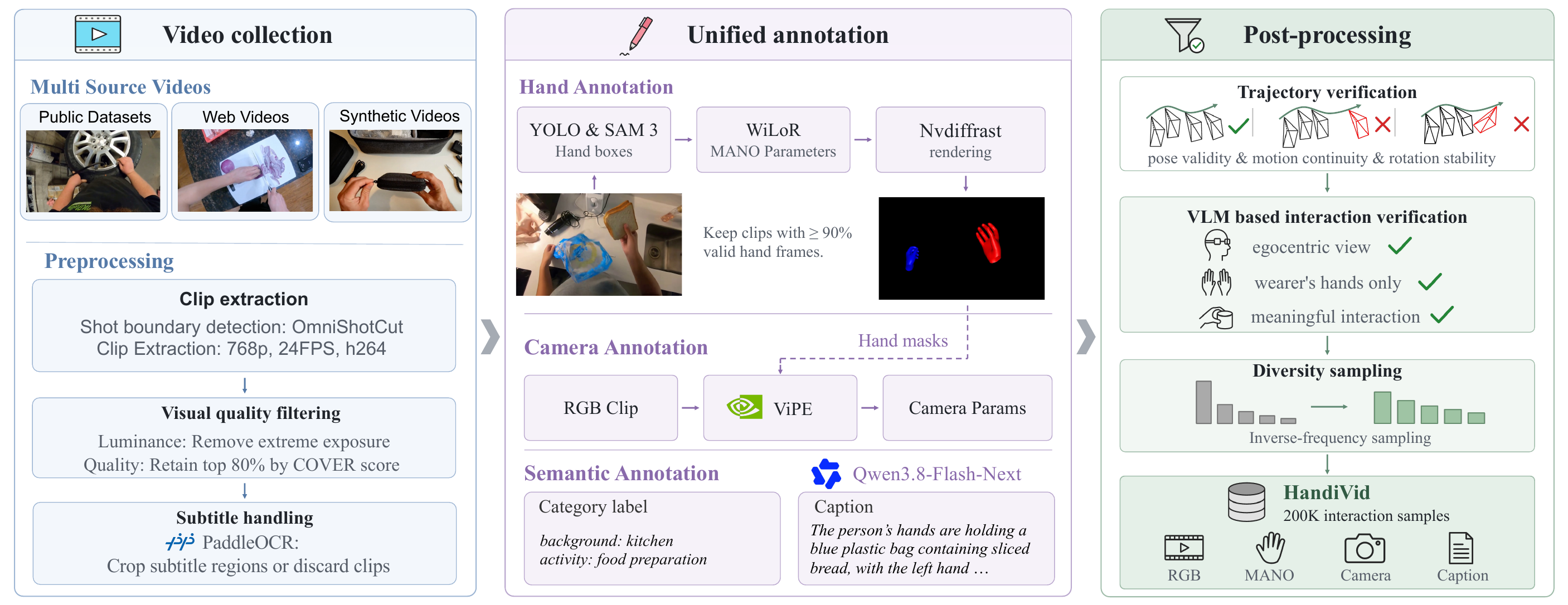}
  \caption{Construction pipeline for \textbf{HandiVid}. Multi-source videos undergo quality filtering, hand and camera annotation, semantic annotation, interaction verification, and diversity sampling.}
  \label{fig:pipeline}
\end{figure}

We construct HandiVid through video collection, preprocessing, unified annotation, and diversity sampling. The resulting dataset contains approximately 200K egocentric interaction clips, each paired with 3D hand parameters, frame-aligned hand renderings, camera trajectories, and interaction descriptions. \Cref{fig:pipeline} summarizes the pipeline.

\subsection{Video collection}
\label{sec:collection}

HandiVid draws on three sources: public datasets, online videos, and egocentric interaction videos synthesized by MiniMax-H3~\cite{minimax2026h3}. Public sources include Open-AoE~\cite{li2026openaoe}, HOI4D~\cite{liu2022hoi4d}, TACO~\cite{liu2024taco}, EgoDex~\cite{hoque2026egodex}, HOT3D~\cite{banerjee2025hot3d}, H2O~\cite{kwon2021h2o}, Ego4D~\cite{grauman2022ego4d}, EPIC-KITCHENS-100~\cite{damen2022rescaling}, CaptainCook4D~\cite{peddi2024captaincook4d}, and ARCTIC~\cite{fan2023arctic}. The collected content covers everyday object manipulation, tool use, and bimanual coordination. For datasets with multiple views, we retain only egocentric recordings. Source videos must have a resolution of at least 720p, a frame rate of at least 24 fps, and more than 200 frames to support subsequent clip extraction and motion annotation. See the appendix for details.

\subsection{Video preprocessing}
\label{sec:preprocessing}

We standardize videos into candidate clips, filter visual quality, and remove subtitles embedded in the image.

\noindent\textbf{Shot boundary detection.}
Online videos often contain edits and transitions that disrupt action and camera continuity. We use OmniShotCut~\cite{wang2026omnishotcut} to locate shot boundaries, retain continuous shots, and exclude dissolves, wipes, and fades. To improve throughput, videos are distributed across GPUs, with batched inference and asynchronous decoding prefetch to overlap decoding and model execution.

\noindent\textbf{Clip extraction and transcoding.}
We extract fixed-length clips within continuous shots. To ensure sharp initial frames for image-to-video generation, we search up to 10 frames after each candidate cut and select the sharpest valid frame based on grayscale Laplacian variance, favoring earlier frames in case of ties. Clips are sampled on a 24 fps timeline using source-aligned timestamps, retaining 125 or 240 frames and discarding incomplete segments. All clips are encoded as H.264 MP4 at 24 fps, with preserved aspect ratios and the shorter side resized to 768 pixels. Further details are provided in the appendix.

\noindent\textbf{Visual quality filtering.}
Before annotation, we filter candidate clips based on luminance and perceptual quality. We first compute RGB-weighted luminance for each sampled frame, excluding a small fraction of the darkest and brightest pixels to mitigate the effects of local shadows and highlights. We then average the remaining luminance values across frames and discard the darkest and brightest 5\% of clips. For the remaining clips, we employ COVER~\cite{he2024cover} to assess technical quality, aesthetic quality, and semantic content, with the semantic branch leveraging CLIP~\cite{radford2021clip} features. The overall quality score is defined as $q_{\mathrm{overall}} = q_{\mathrm{technical}} + q_{\mathrm{aesthetic}} + q_{\mathrm{semantic}}$, where each term represents the mean score of the corresponding branch. We retain the top 80\% of clips based on this score, reducing the computational cost of subsequent hand detection, reconstruction, camera estimation, and semantic annotation.

\noindent\textbf{Subtitle filtering.}
Subtitles may cause generative models to reproduce text overlays unrelated to the interaction. We use PaddleOCR~\cite{du2020ppocr} to detect text in the lower third of each frame and identify subtitles based on detections in five consecutive frames to reduce false positives. We then apply a fixed crop across all frames to remove the subtitle region, retaining clips only if the shorter side exceeds 720 pixels.

\subsection{Video annotation}
\label{sec:annotation}

We generate three complementary annotations: 3D hand motion and renderings, camera parameters, and semantic categories with interaction descriptions.

\noindent\textbf{Hand annotation.}
We first use the hand detector provided by WiLoR~\cite{potamias2025wilor} to locate left and right hands in each frame. Box overlap and temporal consistency checks remove duplicate boxes and transient false positives. A valid hand frame contains at least one retained hand box. Clips with valid detections in fewer than 50\% of frames are discarded.

For the remaining clips, short-gap interpolation and SAM 3~\cite{carion2025sam3} recover missing boxes. For short gaps bracketed by valid detections of the same hand, we linearly interpolate box centers and sizes if the box before the gap is sufficiently far from the image boundary. For remaining gaps, we enlarge the previous box for that hand, run SAM 3 with the prompt \texttt{hand} in the corresponding local region of the current frame, and take the bounding box of the largest segmented instance while retaining the track's left or right identity. We then remove duplicates, interpolate remaining short gaps, and smooth continuous tracks. Clips with valid hand boxes in fewer than 90\% of frames after completion are discarded. SAM 3 is deployed with TensorRT~\cite{tensorrt} and reuses frames decoded during detection, improving annotation coverage without decoding them again.

Given the completed boxes, WiLoR estimates MANO~\cite{romero2017mano} global orientations, articulated poses, shape parameters, and camera-space translations for each frame. We also save the resulting 3D vertices and joints. Parameters are temporally smoothed within continuous valid intervals before mesh generation. Finally, nvdiffrast~\cite{laine2020nvdiffrast} renders the meshes against a black background, producing hand videos with the same frame count, resolution, and frame rate as the RGB clips.

\noindent\textbf{Camera annotation.}
We use NVIDIA ViPE~\cite{huang2025vipe} to estimate per-frame camera poses and intrinsics with metric scale priors. GeoCalib~\cite{veicht2024geocalib} initializes the intrinsics, and MoGe-2~\cite{wang2025moge2} supplies metric monocular depth priors for keyframes during geometric optimization. Let $\widetilde D_k(\mathbf{u})$ denote keyframe depth before scale adjustment, and let $s_k>0$ be the factor predicted by the MoGe-2 scale branch. The depth and inverse-depth priors supplied to the optimizer are
\begin{equation}
  \begin{aligned}
    D_k^{\mathrm{metric}}(\mathbf{u}) &= s_k\widetilde D_k(\mathbf{u}), \\
    d_k^{\mathrm{prior}}(\mathbf{u}) &= \frac{1}{D_k^{\mathrm{metric}}(\mathbf{u})},
  \end{aligned}
  \label{eq:depth}
\end{equation}
where $\mathbf{u}\in\Omega_k^{\mathrm{valid}}$ denotes pixels with finite, positive depth in the model's validity mask. The inverse-depth prior constrains bundle adjustment, placing camera translations and keyframe depths on a shared metric scale.

ViPE originally uses Grounding DINO~\cite{liu2024groundingdino}, SAM~\cite{kirillov2023sam}, and temporal mask propagation to exclude dynamic foreground regions from geometric optimization. Instead, we directly leverage the reconstructed hand renderings to obtain binary hand masks. We apply morphological closing and dilation to these masks and use their complement to define valid regions for camera estimation:
\begin{equation}
  M_t = 1-\operatorname{Dilate}\bigl(\operatorname{Close}(H_t)\bigr).
  \label{eq:handmask}
\end{equation}
where $H_t(\mathbf{u})=1$ denotes hand foreground and $M_t(\mathbf{u})=1$ indicates valid pixels for camera motion estimation. This eliminates the need for additional foreground detection, segmentation, and temporal propagation.

\noindent\textbf{Categories and interaction descriptions.}
We use Qwen3.8-Flash-Next~\cite{qwen2026flashnext}, served through vLLM~\cite{kwon2023vllm}, for category annotation and interaction description generation. Sampled frames include the first frame and are presented in temporal order. For category annotation, we assign scene and activity labels based on visible environments and representative hand-object interactions. Ambiguous cases are labeled as \texttt{unknown}, while activities outside predefined categories are assigned to other activities or other object manipulation. These labels support subsequent content balancing. For interaction descriptions, we generate both long and short versions conditioned on the first frame and subsequent observations. Long descriptions follow the image-to-video prompting format of MiniMax-H3~\cite{minimax2026h3}, detailing the initial scene, hands, objects, and subsequent actions, while short descriptions provide concise interaction summaries. Both versions describe only visually supported actions and state changes, avoiding unsupported details and events between sampled frames. Camera motion is excluded to prevent camera-induced changes from being misinterpreted as hand or object actions.

\subsection{Video postprocessing}
\label{sec:postprocessing}

After annotation, we screen camera trajectories, verify interaction content, and sample clips by semantic category.

\noindent\textbf{Camera outlier removal.}
We reject clips with missing poses, nonfinite values, invalid rotation matrices, or inconsistent numbers of poses and frame indices. We then identify translation jumps relative to local motion, abrupt inter-frame rotations, and sharp turns over short intervals. Displacement directions before and after local windows are also compared to detect repeated large reversals after excluding negligible motion noise.

Let the camera-to-world pose be $T_t=\left[\begin{smallmatrix}R_t&\mathbf{p}_t\\0&1\end{smallmatrix}\right]$. For consecutive frames, define
\begin{equation}
  \begin{aligned}
    \ell_t &= \|\mathbf{p}_{t+1}-\mathbf{p}_t\|_2, \\
    \theta_t &= \arccos\!\left(
      \operatorname{clip}\!\left(
      \frac{\operatorname{tr}(R_t^\top R_{t+1})-1}{2},-1,1
      \right)\right).
  \end{aligned}
  \label{eq:camera}
\end{equation}
Clipping prevents numerical errors from moving the argument outside the domain of $\arccos$. Let $\bar\ell_t$ be the mean displacement in a centered local window, shifted inward at sequence boundaries, containing at most 30 poses and hence 29 displacements. We reject a clip if any $\ell_t>5\bar\ell_t$ or $\theta_t>60^\circ$. We also reject clips containing a relative rotation greater than $100^\circ$ between any two poses separated by at most approximately 1.2 seconds. Rejecting the entire clip when a criterion is triggered reduces discontinuous or excessively abrupt camera conditions.

\noindent\textbf{VLM-based interaction verification.}
We sample one frame every 25 frames and provide the temporally ordered sequence to Qwen3.8-Flash-Next through vLLM. The model verifies three criteria: an egocentric viewpoint, visibility of only the camera wearer's hands without other people, and meaningful hand-object interaction. Clips failing any criterion or receiving uncertain predictions are discarded. Meaningful interaction requires observable manipulation, such as grasping, placing, moving held objects, or using tools. Static hand-object poses, hovering hands, and empty hand gestures are excluded.

\noindent\textbf{Diversity sampling.}
To reduce the dominance of common scenes and frequent activities, we use inverse-frequency sampling based on the scene and activity labels from \cref{sec:annotation}. Each label receives a weight inversely proportional to its frequency in the candidate pool. The two weights are multiplied for each clip and normalized across candidates to form sampling probabilities. Weighted sampling without replacement continues until the target dataset size is reached, increasing the probability of selecting less frequent scenes and activities.

\section{Dataset statistics}
\label{sec:statistics}

We characterize HandiVid through scene and activity coverage, visual quality, viewpoint motion, and text annotations.

% Group the four statistics plots in a full-width 2-by-2 figure.

\begin{figure}[t]
  \centering
  \captionsetup[subfigure]{
    font=normalsize,
    labelformat=parens,
    justification=centering,
    skip=3pt
  }

  % (a) 保持原尺寸
  \begin{subfigure}[t]{0.49\textwidth}
    \vspace{0pt}
    \centering
    \includegraphics[width=\linewidth]{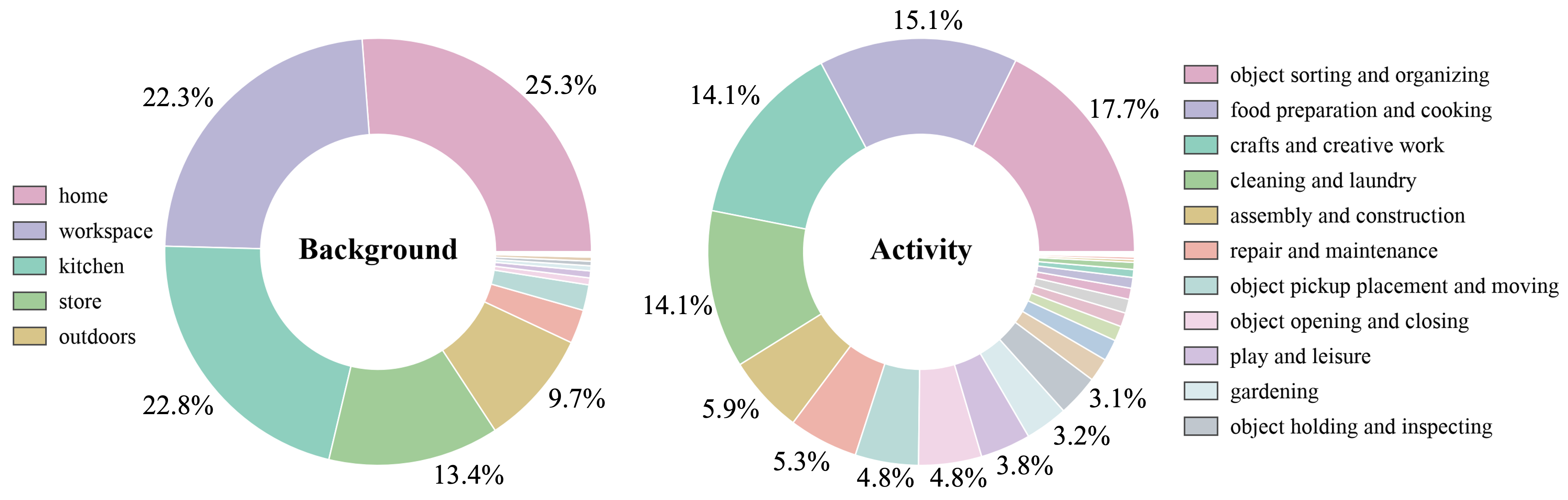}
    \caption{Scene and activity distributions}
    \label{fig:categories}
  \end{subfigure}\hfill
  % (b) 缩小
  \begin{subfigure}[t]{0.49\textwidth}
    \vspace{0pt}
    \centering
    \includegraphics[width=0.8\linewidth]{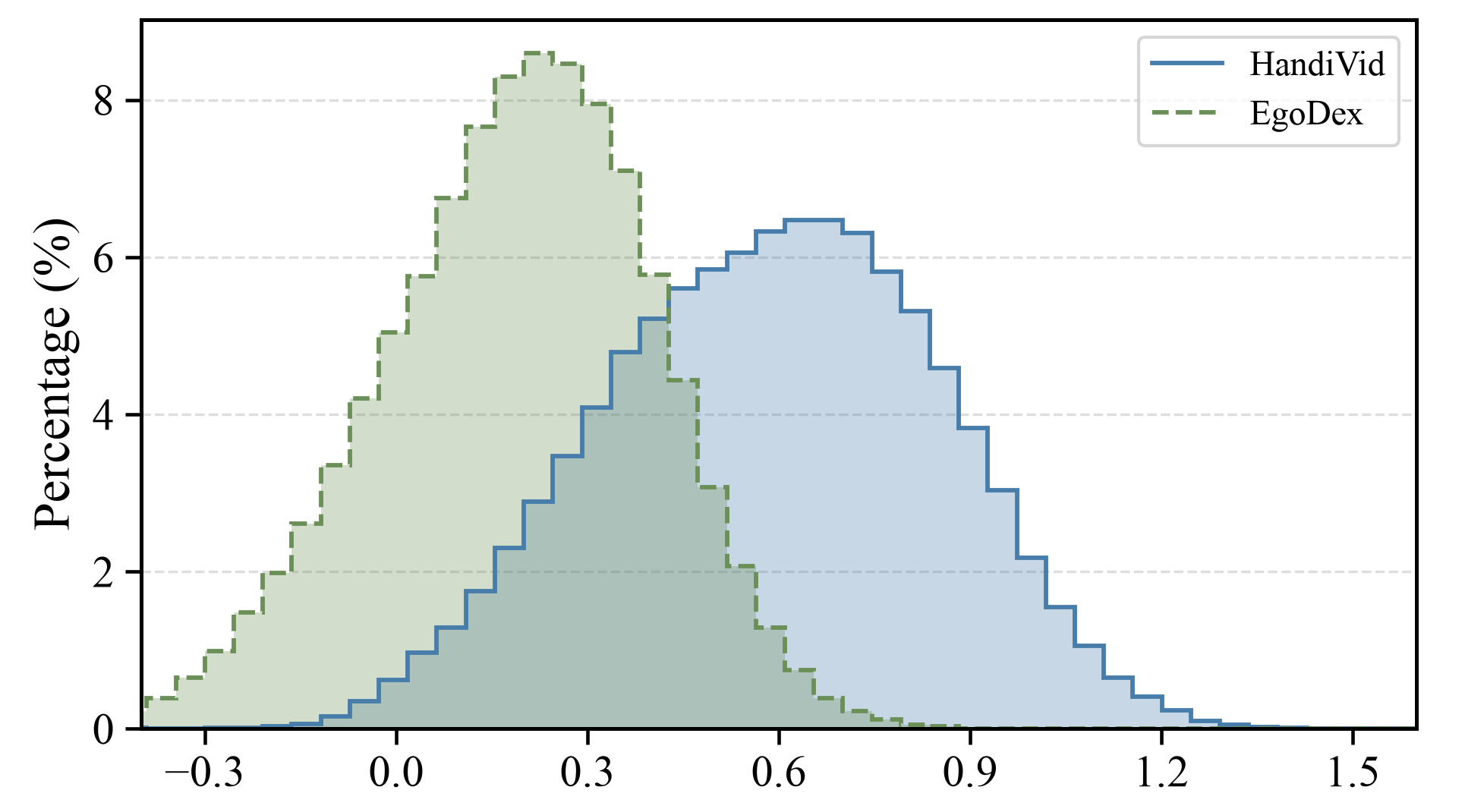}
    \caption{Overall video quality score}
    \label{fig:quality}
  \end{subfigure}

  \par\vspace{2pt}

  % (c) 缩小
  \begin{subfigure}[b]{0.49\textwidth}
    \centering
    \includegraphics[width=0.8\linewidth]{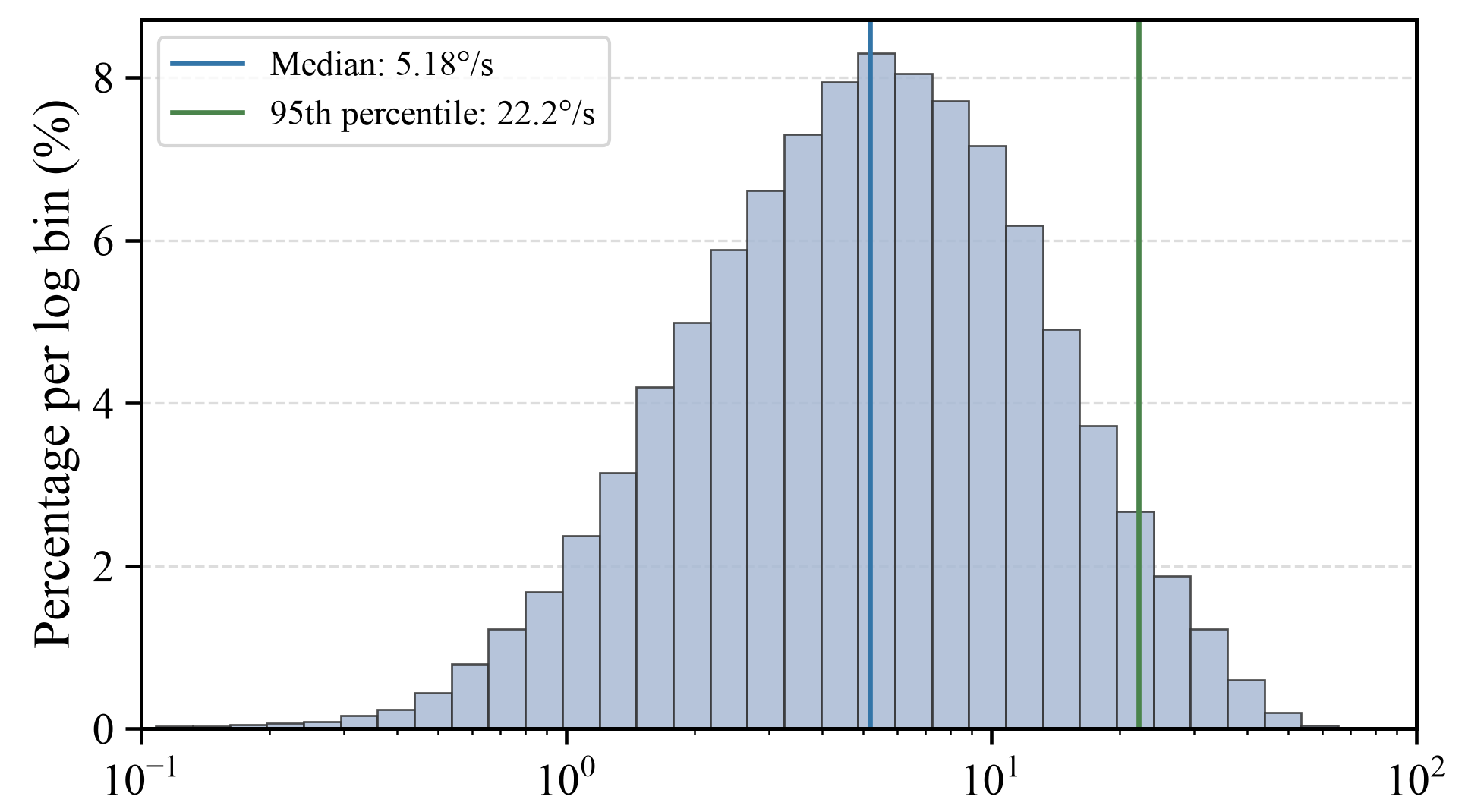}
    \caption{Mean camera angular speed ($^\circ$/s)}
    \label{fig:camera}
  \end{subfigure}\hfill
  % (d) 缩小
  \begin{subfigure}[b]{0.49\textwidth}
    \centering
    \includegraphics[width=0.8\linewidth]{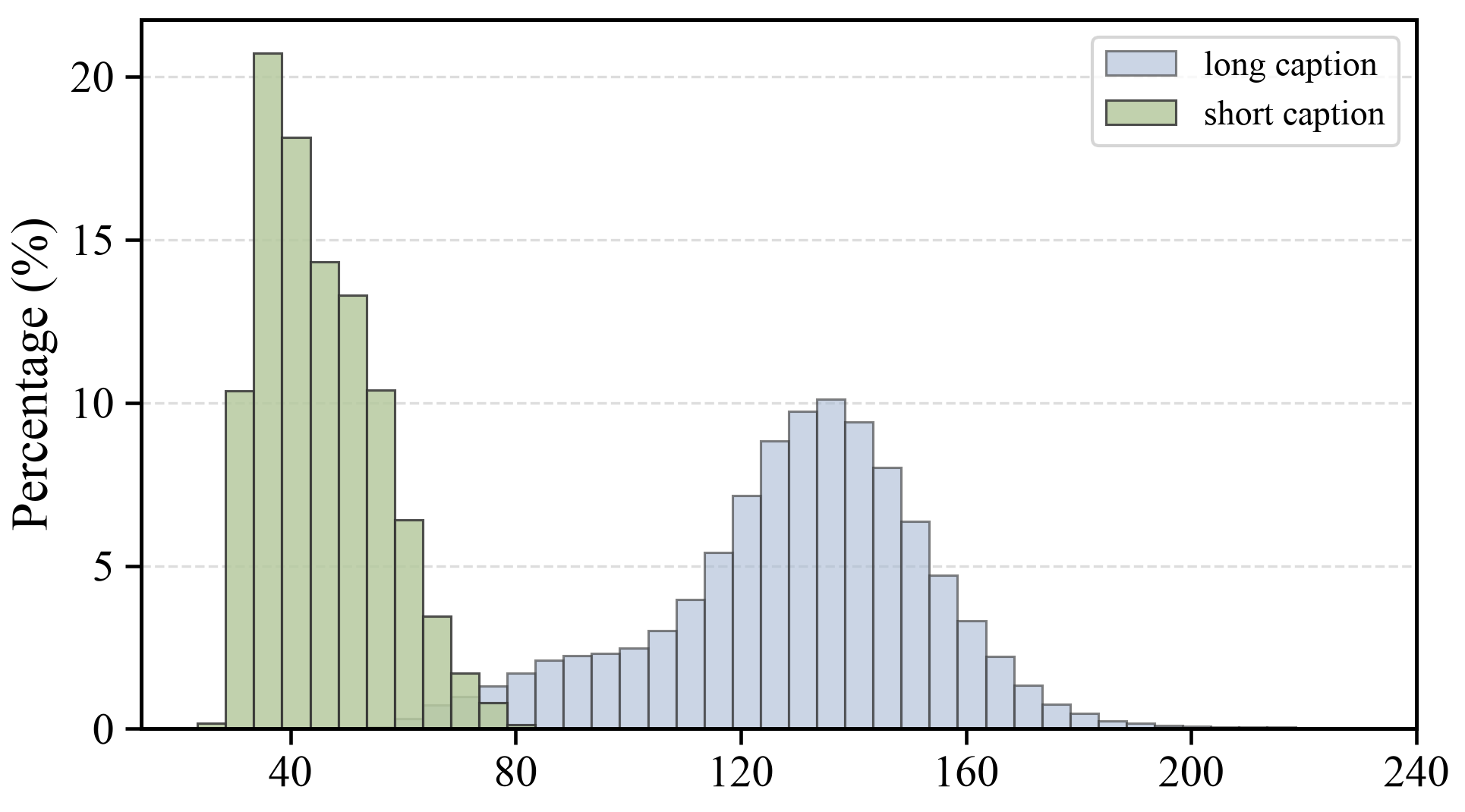}
    \caption{Caption length (words)}
    \label{fig:captions}
  \end{subfigure}

  \caption{Statistics of the proposed \textbf{HandiVid} dataset.}
  \label{fig:statistics}
\end{figure}

\noindent\textbf{Scenes and interactions.}
As shown in \cref{fig:categories}, HandiVid covers homes, kitchens, workspaces, stores, and outdoor environments. Homes, kitchens, and workspaces account for 25.3\%, 22.8\%, and 22.3\% of clips, respectively, while stores and outdoor settings together contribute approximately 23.1\%. These environments provide varied background layouts and manipulation contexts.

Activities include basic actions such as picking up, placing, opening, and closing objects, as well as cooking, crafts, cleaning, assembly, and repair. Object sorting and organizing is the largest category at 17.7\%. Food preparation and cooking, crafts and creative work, and cleaning and laundry account for 15.1\%, 14.1\%, and 14.1\% of clips, respectively. The remaining clips cover assembly, repair, gardening, and other activities, providing interactions with different operational goals.

\noindent\textbf{Visual quality.}
\Cref{fig:quality} compares the overall COVER scores of HandiVid and EgoDex, computed as the sum of technical, aesthetic, and semantic quality scores. HandiVid exhibits a clear shift toward higher scores, with most clips concentrated between 0.3 and 0.9, indicating improved perceptual quality compared to EgoDex.

\noindent\textbf{Viewpoint motion.}
HandiVid captures diverse viewpoint motions during egocentric manipulation. For each clip, we measure mean camera angular speed as accumulated rotation divided by clip duration. Across clips, the median is $5.18^\circ/\mathrm{s}$ and the 95th percentile is $22.2^\circ/\mathrm{s}$ (\cref{fig:camera}). These statistics highlight the wide range of camera motions captured in HandiVid, spanning both relatively stable viewpoints and rapid viewpoint changes during manipulation.

\noindent\textbf{Text annotations.}
Each clip includes a long description detailing the initial scene and subsequent actions, and a short description summarizing the overall interaction. Short descriptions typically contain 30 to 60 English words, while long descriptions mostly range from 110 to 160 words (\cref{fig:captions}). Both capture visible objects, actions, and temporal progression at different levels of detail.

\begin{table}[t]
  \caption{Generation quality and control consistency across HandiVid training checkpoints on the fixed 100-clip evaluation set. Evaluation scenes are absent from training. W.~CLIP and MANO~D.\ denote weighted CLIP similarity and MANO Dice, respectively. Arrows indicate the preferred direction; best values are bold.}
  \label{tab:checkpoints}
  \centering
  \resulttablestyle
  \begin{tabular*}{\textwidth}{@{\extracolsep{\fill}}l*{9}{c}@{}}
    \toprule
    Step & FVD $\downarrow$ & PSNR $\uparrow$ & LPIPS $\downarrow$ & SSIM $\uparrow$ & SC $\uparrow$ & BC $\uparrow$ & W.~CLIP $\uparrow$ & MANO~D. $\uparrow$ & Cam-BG $\uparrow$ \\
    \midrule
    50    & 521.25 & 19.39 & 0.2467 & 0.7324 & 94.73 & 94.27 & 83.44 & 85.61 & 56.26 \\
    500   & 513.87 & 19.54 & 0.2417 & 0.7352 & 94.77 & 94.29 & 83.49 & 85.93 & 57.53 \\
    1,000 & 486.24 & 20.11 & 0.2265 & 0.7419 & 94.87 & 94.32 & 83.56 & 88.81 & 60.69 \\
    1,500 & 414.41 & 20.89 & 0.2106 & 0.7503 & 95.11 & 94.41 & 83.83 & 92.57 & 64.12 \\
    2,000 & \textbf{409.63} & \textbf{21.35} & \textbf{0.2006} & \textbf{0.7644} & \textbf{95.18} & \textbf{94.72} & \textbf{83.86} & \textbf{92.64} & \textbf{64.75} \\
    \bottomrule
  \end{tabular*}
\end{table}

\begin{table}[t]
  \caption{Comparison of training data on the fixed 200-clip evaluation set. Both models use the same evaluation scenes, which are absent from training. W.~CLIP and MANO~D.\ denote weighted CLIP similarity and MANO Dice, respectively. Better values are bold.}
  \label{tab:trainingdata}
  \centering
  \resulttablestyle
  \begin{tabular*}{\textwidth}{@{\extracolsep{\fill}}l*{9}{c}@{}}
    \toprule
    Model & FVD $\downarrow$ & PSNR $\uparrow$ & LPIPS $\downarrow$ & SSIM $\uparrow$ & SC $\uparrow$ & BC $\uparrow$ & W.~CLIP $\uparrow$ & MANO~D. $\uparrow$ & Cam-BG $\uparrow$ \\
    \midrule
    ARCTIC   & 554.81 & 18.14 & 0.2850 & 0.6947 & 94.16 & 93.54 & 82.73 & 87.06 & 48.10 \\
    HandiVid & \textbf{358.55} & \textbf{21.21} & \textbf{0.2057} & \textbf{0.7451} & \textbf{94.85} & \textbf{94.46} & \textbf{83.61} & \textbf{92.74} & \textbf{64.91} \\
    \bottomrule
  \end{tabular*}
\end{table}

\begin{table}[t]
  \caption{Cross-dataset evaluation on H2O and TACO, with 20 fixed clips per dataset. Clips from both datasets are excluded from the HandiVid training split. W.~CLIP and MANO~D.\ denote weighted CLIP similarity and MANO Dice, respectively. Better values within each dataset are bold.}
  \label{tab:crossdataset}
  \centering
  \resulttablestyle
  \begin{tabular*}{\textwidth}{@{\extracolsep{\fill}}ll*{9}{c}@{}}
    \toprule
    Dataset & Model & FVD $\downarrow$ & PSNR $\uparrow$ & LPIPS $\downarrow$ & SSIM $\uparrow$ & SC $\uparrow$ & BC $\uparrow$ & W.~CLIP $\uparrow$ & MANO~D. $\uparrow$ & Cam-BG $\uparrow$ \\
    \midrule
    \multirow{2}{*}{H2O}
      & ARCTIC   & 1051.83 & 18.79 & 0.2556 & 0.6952 & 94.40 & 93.59 & 82.81 & 91.79 & 50.23 \\
      & HandiVid & \textbf{670.55} & \textbf{21.13} & \textbf{0.1748} & \textbf{0.7335} & \textbf{94.84} & \textbf{94.67} & \textbf{83.24} & \textbf{93.75} & \textbf{59.09} \\
    \midrule
    \multirow{2}{*}{TACO}
      & ARCTIC   & 1024.00 & 18.20 & 0.3686 & 0.5754 & 92.17 & 92.14 & 81.17 & 91.44 & 49.53 \\
      & HandiVid & \textbf{532.31} & \textbf{20.25} & \textbf{0.2773} & \textbf{0.6024} & \textbf{92.20} & \textbf{92.18} & \textbf{82.04} & \textbf{92.51} & \textbf{71.96} \\
    \bottomrule
  \end{tabular*}
\end{table}

\begin{figure}[t]
  \centering
  \includegraphics[width=\textwidth]{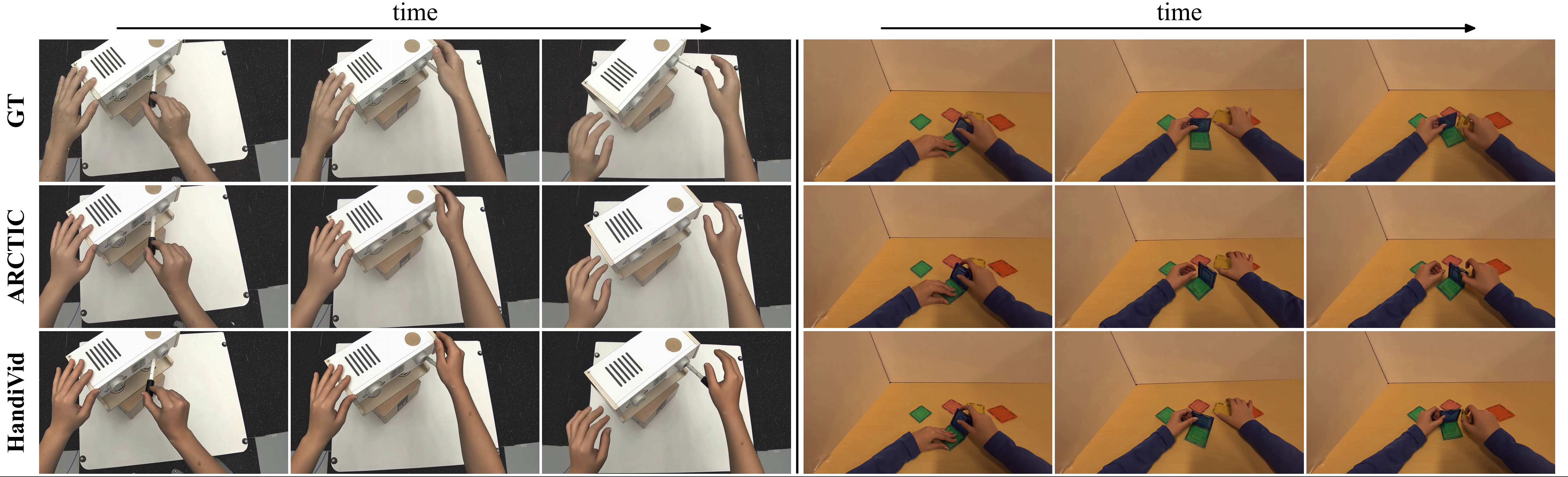}
  \caption{\textbf{Interaction generation with different training data.} The left example shows tool use and the right shows bimanual manipulation of colored components. Rows show ground truth, the ARCTIC model, and the HandiVid model. Columns within each example show three corresponding times. Both models receive identical initial images and control conditions.}
  \label{fig:comparison}
\end{figure}

\begin{figure}[t]
  \centering
  \includegraphics[width=\textwidth]{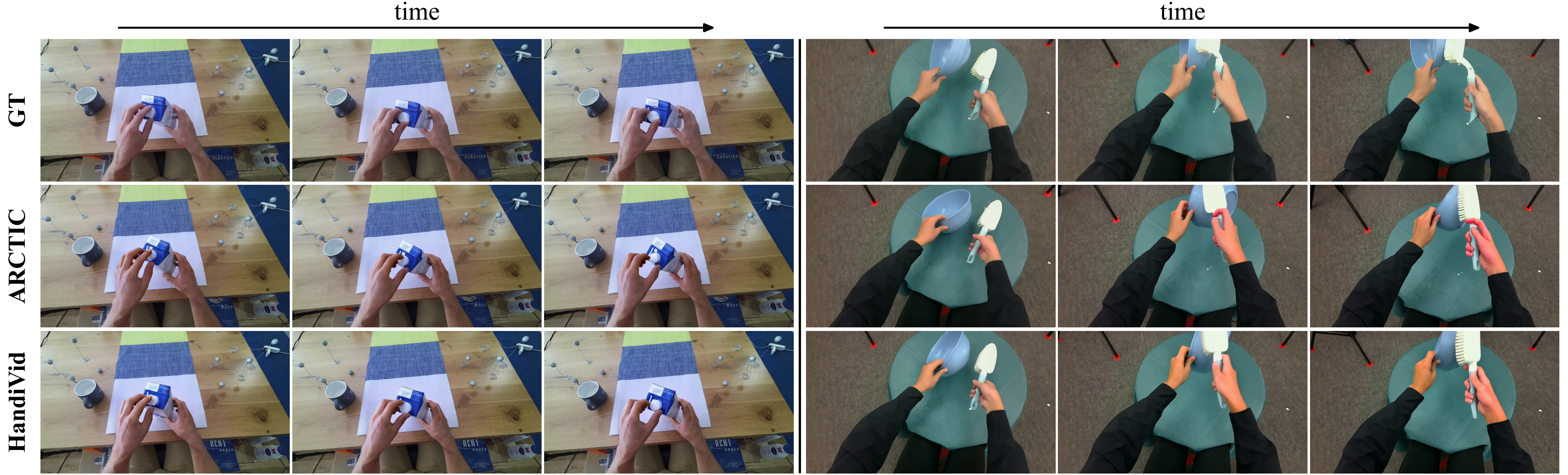}
  \caption{\textbf{Cross-dataset generation on H2O and TACO.} The left example shows opening a carton cap in H2O; the right shows manipulating a container and brush in TACO. Rows show ground truth, the ARCTIC model, and the HandiVid model. Columns within each example show corresponding times in chronological order.}
  \label{fig:ood}
\end{figure}

\section{Experiments}
\label{sec:experiments}

We first assess category labels, MANO annotations, and interaction validity through manual inspection, and quantify hand box completion. We then evaluate the dataset through hand-driven egocentric video generation.
\subsection{Annotation quality}
\label{sec:annotationquality}

We randomly sample 500 clips spanning diverse scenes and activities. Coauthors manually assess category labels, MANO annotations, and interaction validity.

\noindent\textbf{Category labels.}
We compare the predicted scene and activity labels with manual annotations for each of the 500 sampled clips. A clip passes only if both labels agree with the manual annotations. Over 92\% of clips meet this joint criterion.

\noindent\textbf{Hand annotations and interaction validity.}
We assess MANO quality by overlaying rendered meshes on RGB frames. Clips with misalignment, missed hands, false detections, or incorrect handedness are considered failures. The MANO pass rate reaches 95\%, and all 500 clips contain valid interactions as defined in \cref{sec:postprocessing}.

\noindent\textbf{Hand box completion.}
Among successfully processed videos, 13.07\% contain valid hand boxes recovered by SAM 3. Within these videos, SAM 3 recovers left and right hand boxes in 3.74\% and 3.19\% of frames, respectively, corresponding to 0.49\% and 0.42\% across all analyzed inputs. These boxes support 3D hand reconstruction.

\subsection{Interactive video generation}
\label{sec:setup}

\noindent\textbf{Implementation details.}
We build a full-sequence video generation baseline on pretrained MiniMax-H3, conditioned on hand motion, camera poses and captions. Given an initial frame, an interaction description, hand renderings, and camera poses, the model generates egocentric interaction videos. The VAE encodes hand renderings into latent tokens, which are combined with image, text, and noisy video tokens. Camera intrinsics and extrinsics relative to the first frame are incorporated via PRoPE~\cite{li2025prope}. For training, we extract 124-frame sequences from the retained clips and resize them with their aspect ratios preserved to a shorter side of 768 pixels.

\noindent\textbf{Training data and evaluation sets.}
We compare two models based on the pretrained backbone: one trained using only ARCTIC clips and the other using HandiVid. In the generation experiments, we refer to these models as ARCTIC and HandiVid, respectively. ARCTIC is also one of the source datasets of HandiVid. H2O and TACO clips are excluded from the HandiVid training split to assess cross-dataset generalization. We use 100 fixed, manually selected clips for checkpoint evaluation and a separate set of 200 clips for training-data comparison. Cross-dataset evaluation additionally uses 20 fixed clips from each of H2O and TACO. All evaluation scenes are disjoint from training.

\noindent\textbf{Evaluation metrics.}
We evaluate generation quality using FVD, PSNR, LPIPS~\cite{zhang2018lpips}, SSIM~\cite{wang2004ssim}, and weighted CLIP similarity. Temporal consistency is assessed using subject consistency (SC) and background consistency (BC) from VBench~\cite{huang2024vbench}. MANO Dice measures projected hand silhouette overlap, while Cam-BG evaluates background motion agreement with reference videos. Detailed metric definitions are in the appendix.

\subsection{Quantitative results}
\label{sec:quantitative}

\Cref{tab:checkpoints} reports the performance of the HandiVid model across training checkpoints, with step 2,000 achieving the best results on all nine metrics. From step 50 to 2,000, FVD decreases from 521.25 to 409.63, PSNR increases by 1.96 dB, and LPIPS decreases from 0.2467 to 0.2006. Meanwhile, MANO Dice and Cam-BG improve by 7.03 and 8.49 points, respectively. These results indicate progressive improvements in both visual quality and control consistency during training.

We further investigate the impact of training data by comparing ARCTIC and HandiVid on a fixed set of 200 clips, separate from the checkpoint evaluation set. As shown in \cref{tab:trainingdata}, HandiVid outperforms ARCTIC across all nine metrics. Specifically, FVD decreases from 554.81 to 358.55, PSNR improves by 3.07 dB, and LPIPS decreases from 0.2850 to 0.2057. MANO Dice and Cam-BG also increase by 5.68 and 16.81 points, respectively. These improvements demonstrate the effectiveness of HandiVid in enhancing generation quality, hand projection consistency, and background motion consistency.

To evaluate out-of-distribution (OOD) generalization, we test both models on H2O and TACO, using 20 clips from each dataset. Clips from neither dataset are included in the HandiVid training split. As reported in \cref{tab:crossdataset}, HandiVid consistently outperforms ARCTIC across all nine metrics on both datasets. On H2O, FVD decreases from 1051.83 to 670.55, PSNR improves by 2.34 dB, and Cam-BG increases by 8.86 points. On TACO, FVD decreases from 1024.00 to 532.31, with a 2.05 dB improvement in PSNR and a 22.43-point gain in Cam-BG. These results suggest that the benefits of HandiVid extend to unseen data sources, particularly in visual quality and background motion consistency.

\subsection{Qualitative results}
\label{sec:qualitative}

We compare ARCTIC and HandiVid under identical initial frame, text, hand rendering, and camera conditions. \Cref{fig:comparison} presents qualitative comparisons on the original evaluation domain. Both models generate scenes consistent with the input conditions, but HandiVid better preserves fine object details and spatial relationships. In the tool-use example, ARCTIC loses tool details and exhibits deviations in relative hand and object positions in later frames, whereas HandiVid more closely follows the reference configurations. Similarly, in the colored-component example, HandiVid produces more accurate grips and relative component positions, demonstrating improved consistency in hand and object interactions.

We further examine OOD generation on H2O and TACO in \cref{fig:ood}. In the H2O carton-opening example, ARCTIC exhibits deviations in finger placement around the cap, while HandiVid generates finger configurations closer to the reference. In the TACO container-and-brush example, HandiVid better preserves the index finger position and grip compared with ARCTIC. These observations are consistent with the quantitative improvements in hand projection consistency on the OOD datasets.

\section{Conclusion}
\label{sec:conclusion}

We introduce HandiVid, a high-quality dataset of approximately 200K egocentric interaction clips with aligned 3D hand parameters, hand renderings, camera trajectories, and interaction descriptions. A unified pipeline combines quality filtering, hand detection completion, interaction verification, and diversity sampling to curate clips from public datasets, online videos, and model-generated content. Experiments show that training on HandiVid improves generation quality and motion consistency over training on ARCTIC, with gains extending to unseen data sources. Remaining annotation errors motivate future refinement of hand reconstruction and camera estimation, particularly under occlusion and rapid motion. We will release the dataset and annotation pipeline to support hand-driven interactive video generation.
\appendix
\setcounter{table}{0}
\setcounter{figure}{0}
\setcounter{equation}{0}
\renewcommand{\thetable}{S\arabic{table}}
\renewcommand{\thefigure}{S\arabic{figure}}
\renewcommand{\theequation}{S\arabic{equation}}
\section*{Appendix}

\section{Video sources}
\label{app:sources}

HandiVid combines public datasets, online videos, and videos synthesized by MiniMax-H3. Public datasets provide natural activities and detailed manipulation recordings, online videos broaden the range of everyday scenes and recording styles, and synthetic videos supplement interaction content organized by category. All sources undergo the filtering, clip extraction, and unified annotation pipeline described in the main paper. \Cref{tab:sources} reports the final composition after filtering and annotation.

\begin{table}[H]
  \caption{Final composition of HandiVid. Counts refer to retained clips after filtering and annotation.}
  \label{tab:sources}
  \centering
  \begin{tabular}{lrr}
    \toprule
    Source & Clips & Proportion \\
    \midrule
    Public datasets & 160K & 80\% \\
    Online videos & 20K & 10\% \\
    MiniMax-H3 synthetic videos & 20K & 10\% \\
    \midrule
    Total & 200K & 100\% \\
    \bottomrule
  \end{tabular}
\end{table}

\subsection{Open-AoE}

Open-AoE~\cite{li2026openaoe} records egocentric manipulation in natural environments using consumer smartphones. It provides videos, MANO hand poses, camera trajectories, and temporally bounded atomic action descriptions for embodied learning. Both the original sequences and action intervals vary in length. HandiVid reorganizes these recordings into fixed-length generation clips, generates descriptions from the visible actions after cropping, and supplies aligned hand renderings and camera conditions so that text and motion correspond to the same interval.

\subsection{TACO}

TACO~\cite{liu2024taco} captures bimanual manipulation involving combinations of tools, actions, and objects, including the use of tools to manipulate other objects. It provides synchronized first-person and third-person videos, detailed 3D hand-object meshes, and action labels for tasks such as recognition, motion prediction, and collaborative grasping. HandiVid selects egocentric tool-use and bimanual clips, standardizes their format, and describes the visible process so that they can be used jointly with other sources for generation.

\subsection{EgoDex}

EgoDex~\cite{hoque2026egodex} uses Apple Vision Pro to collect 829 hours of dexterous tabletop manipulation across 194 tasks. Original annotations include 3D head, upper-limb, and hand poses, camera intrinsics and poses, and natural language. HandiVid extracts generation clips, regenerates descriptions for the cropped content, and constructs a common set of MANO parameters, hand renderings, and camera conditions for training across recording devices and sources.

\subsection{HOT3D}

HOT3D~\cite{banerjee2025hot3d} supports egocentric 3D hand-object tracking with synchronized multiview recordings from head-mounted devices. It provides accurate 3D hand and object poses, hand and object models, MANO and UmeTrack hand annotations, and short clips for tracking evaluation. HandiVid selects egocentric manipulation videos by interaction content and visual quality, then generates descriptions and organizes common generation conditions.

\subsection{H2O}

H2O~\cite{kwon2021h2o} records two hands manipulating everyday objects for egocentric interaction recognition. It provides synchronized multiview RGB-D videos, 3D left and right hand poses, 6D object poses, camera poses, object models, and interaction labels. HandiVid selects the egocentric view, standardizes clips and motion conditions, and describes the manipulated objects, actions, and temporal order for joint hand and camera control.

\subsection{HOI4D}

HOI4D~\cite{liu2022hoi4d} provides egocentric RGB-D videos with hand-object interaction annotations, object poses, action segmentation, and hand poses. HandiVid selects suitable egocentric manipulation clips and applies the same clip extraction and unified annotation pipeline.

\subsection{Ego4D}

Ego4D~\cite{grauman2022ego4d} records daily activities across diverse environments, participants, and interaction contexts. It provides narrations and annotations for tasks such as episodic memory, hand-object interaction, and forecasting, but does not uniformly provide per-frame MANO parameters and camera trajectories for every video. HandiVid extracts continuous clips with purposeful manipulation, removes low-quality and noninteractive content, and supplies unified motion annotations and clip-specific descriptions.

\subsection{EPIC-KITCHENS-100}

EPIC-KITCHENS-100~\cite{damen2022rescaling} captures unscripted activities in participants' kitchens, including food preparation, cooking, cleaning, and object manipulation. It provides temporally bounded action narrations, verb and noun labels, and automatic annotations such as hand-object detections and object masks. HandiVid extracts manipulation clips of the required format, generates long and short descriptions from the visible content, and supplies 3D hand parameters, renderings, and camera trajectories.

\subsection{CaptainCook4D}

CaptainCook4D~\cite{peddi2024captaincook4d} records recipe execution in real kitchens, including both normal procedures and execution errors. It provides step boundaries, error categories and descriptions, and fine-grained action annotations for a subset of recordings, supporting error recognition, step localization, and procedural learning. HandiVid extracts short manipulation intervals from the longer records and provides descriptions and motion annotations corresponding to the cropped content.

\subsection{ARCTIC}

ARCTIC~\cite{fan2023arctic} captures bimanual manipulation of articulated objects, such as opening books, using scissors, and manipulating laptops. It provides accurate 3D hand and object meshes, object motion, and dynamic contact information for consistent motion reconstruction and interaction field estimation. HandiVid selects its egocentric videos and organizes them into generation clips with descriptions and common motion conditions for joint training with other sources.

\subsection{Online videos}

We first manually collect links to high-quality egocentric manipulation videos from popular YouTube creators, then expand the search using keywords such as egocentric and first-person. These videos generally lack standardized 3D hand annotations, camera trajectories, and clip-specific descriptions. Shot segmentation, visual quality filtering, interaction verification, and unified annotation yield 20K retained clips.

\subsection{Synthetic videos}

We use DeepSeek-V4-Pro~\cite{deepseek2026v4pro} to generate interaction categories, then produce batches of captions describing egocentric hand-object manipulation for each category. These captions prompt MiniMax-H3 to synthesize videos by category. The synthetic videos undergo the same filtering and annotation pipeline as the real videos, yielding 20K retained clips. The synthesis captions serve as generation prompts; the final released interaction descriptions are produced by the unified semantic annotation process from the content actually visible in the generated videos.

\section{Clip start selection}
\label{app:clipstart}

Let $\mathcal{C}$ be the nonempty set of at most 10 candidate start frames beginning at the current cut position. Every candidate must leave enough frames for the target clip length. If no candidate satisfies this requirement, the remaining tail is discarded. Let $G_t$ be the grayscale image at frame $t$ after resizing with the aspect ratio preserved. We select
\begin{equation}
  t^\star = \underset{t\in\mathcal{C}}{\arg\max}\;
  \operatorname{Var}_{\mathbf{u}\in\Omega^\circ}
  \bigl[\Delta G_t(\mathbf{u})\bigr],
  \label{eq:clipstart}
\end{equation}
where $\Delta$ is the discrete Laplacian and $\Omega^\circ$ excludes the outermost pixel boundary. Variance is computed over spatial pixels, and ties are resolved by selecting the earlier candidate.

\section{Metrics and evaluation protocol}
\label{app:metrics}

Evaluation clips are manually selected and fixed, with scenes absent from training. Checkpoint evaluation uses 100 clips, training-data comparison uses a separate set of 200 clips, and cross-dataset evaluation uses 20 clips each from H2O and TACO. Each metric uses the valid samples defined below, with coverage recorded. PSNR is reported in dB. SC, BC, weighted CLIP, MANO Dice, and Cam-BG are reported after multiplying their underlying similarity or consistency values by 100. SSIM and LPIPS retain their original scales.

\subsection{Frame alignment and aggregation}
\label{app:alignment}

Paired metrics align generated and reference frames using recorded source frame indices without temporal interpolation. PSNR and SSIM resize the complete image with its aspect ratio preserved and center-pad it to $1344\times768$, including padding in the score. LPIPS, CLIP, FVD, and Cam-BG use aligned $640\times360$ images followed by their respective preprocessing. SC and BC use complete generated sequences with fixed feature transforms. MANO Dice retains native geometric projection coordinates. Evaluation includes the complete sequence without additionally excluding the conditioning first frame; metric-specific sampling is described below.

For scalar metrics computed per video, we first compute the video score, average within each source, and then average sources equally:
\begin{equation}
  \bar s = \frac{1}{|\mathcal{S}|}
  \sum_{d\in\mathcal{S}}
  \frac{1}{|\mathcal{V}_d|}
  \sum_{v\in\mathcal{V}_d}s_v,
  \label{eq:aggregation}
\end{equation}
where $\mathcal{V}_d$ is the common set of valid videos from source $d$ for that metric, and $\mathcal{S}$ includes only sources with nonempty valid sets. A metric is not reported if the entire evaluation set has no valid videos. FVD is computed over features from the entire evaluation set, rather than averaging source-specific FVD values. Video identities, input conditions, frame mappings, and valid sets are fixed across checkpoints. Failed generations are not replaced by other samples. Reconstruction and motion-estimation failures are handled as specified for MANO Dice and Cam-BG.

\subsection{Standard visual metrics}

FVD, PSNR, LPIPS, and SSIM use their standard definitions. FVD uniformly samples 32 frames per video, extracts 400-dimensional features with a frozen Kinetics-I3D encoder~\cite{carreira2017i3d}, and computes a single score over the evaluation set. PSNR first pools RGB mean squared error over all paired frames in a video and then computes the video-level score. SSIM operates on grayscale images with an $11\times11$ Gaussian window of standard deviation 1.5, removes the outer five pixels of the similarity map, and averages frame scores within each video. LPIPS uses a frozen VGG network~\cite{simonyan2015vgg}, resizes images to $448\times256$, maps pixels to $[-1,1]$, and averages within each video. All models use identical preprocessing and evaluation samples.

\subsection{Weighted CLIP}

We construct a composite similarity score using a fixed OpenAI CLIP ViT-B/32 encoder. For $T>1$ evaluation frames, let $u_t$, $v_t$, and $q$ be unit-normalized features of the generated frame, synchronized reference frame, and shared description, respectively. We compute
\begin{equation}
  \begin{aligned}
    C_T &= \frac{100}{T}\sum_{t=0}^{T-1}u_t^\top q, \\
    C_F &= \frac{100}{T-1}\sum_{t=1}^{T-1}u_t^\top u_{t-1}, \\
    C_V &= \frac{100}{T}\sum_{t=0}^{T-1}u_t^\top v_t, \\
    C_{\mathrm{CLIP}} &= 0.2C_T+0.2C_F+0.6C_V.
  \end{aligned}
  \label{eq:clip}
\end{equation}
The components measure text alignment, adjacent-frame consistency, and reference-video similarity. Cosine values are not clipped at zero. Images pass through the frozen processor with a 224-pixel center crop. All checkpoints use the same descriptions after removing metadata wrappers, with truncation at 77 tokens.

\subsection{Subject and background consistency}

SC and BC follow the frozen VBench implementation and use DINO ViT-B/16~\cite{caron2021dino} and CLIP ViT-B/32 features, respectively. Each generated frame is compared with the first and previous frames using nonnegative cosine similarity, and scores are averaged within the video. SC resizes the shorter side to 224 pixels and uses ImageNet normalization without center cropping. BC resizes the shorter side to 224 pixels, applies a center crop, and uses CLIP normalization. Both metrics use all generated frames without reference videos. Video-level scores are aggregated as in \cref{app:alignment}.

\subsection{MANO Dice}

Frozen WiLoR independently detects and reconstructs hands from generated videos, without access to the driving hand video or reference MANO parameters. We project all triangular mesh faces onto a native canvas of height 768 and take the union of the two hand silhouettes, denoted $M_t^R$ and $M_t^G$ for reference and generated hands. Let $\mathcal{I}_v$ be the frame indices selected for video $v$ at 4 Hz as described below, and define $\mathcal{T}_v=\{t\in\mathcal{I}_v:|M_t^R|>0\}$. For videos with $|\mathcal{T}_v|>0$, we compute
\begin{equation}
  D_{\mathrm{MANO}} =
  \frac{100}{|\mathcal{T}_v|}
  \sum_{t\in\mathcal{T}_v}
  \frac{2|M_t^R\cap M_t^G|}{|M_t^R|+|M_t^G|}.
  \label{eq:mano}
\end{equation}
Target times are spaced at 4 Hz. For each time, the nearest source frame common to all compared outputs is selected, and duplicates are removed. We apply no temporal interpolation, pose alignment, or mask dilation. A missing generated hand receives zero when the reference silhouette is nonempty. Additional generated regions remain in the denominator. Frames with empty reference silhouettes are excluded; videos without any valid reference frames are omitted, with coverage recorded. Silhouette unions retain differences in position, scale, and shape, but do not distinguish left and right identities or directly measure occlusion and physical contact.

WiLoR uses a detection confidence threshold of 0.6, retaining the highest-confidence box for each hand. Boxes are expanded by 1.2, and the reconstruction crop factor is 1.5. Each reference hand uses all 778 vertices, preserves left and right conventions, and applies camera translation exactly once. For width $W$ and height $H$, the projection focal length is $f=(5000/256)\max(W,H)$ and the principal point is $(W/2,H/2)$, followed by the recorded spatial mapping. Reference fits may include five-frame smoothing; generated reconstructions are not smoothed.

\subsection{Cam-BG}

We evaluate background motion at 4 Hz and $640\times360$ resolution. For a video with $T$ frames and source frame rate $f_{\mathrm{src}}$, the sampled indices are
\begin{equation}
  \begin{aligned}
    i_j &= \left\lfloor\frac{j f_{\mathrm{src}}}{4}+\frac{1}{2}\right\rfloor, \\
    j &= 0,\ldots,\left\lfloor\frac{4(T-1)}{f_{\mathrm{src}}}\right\rfloor.
  \end{aligned}
  \label{eq:cambgsampling}
\end{equation}
For each adjacent pair of sampled frames, we take the union of the hand-conditioning foregrounds, compute its convex hull, and dilate it with a $31\times31$ elliptical kernel. We exclude this region, a 12-pixel image border, and persistent near-black padding in the reference. This exclusion depends only on reference conditions, preventing selection of evaluation regions based on generated quality.

We estimate background displacement using Shi-Tomasi corners~\cite{shi1994features}, pyramidal Lucas-Kanade tracking~\cite{lucas1981registration,bouguet2001pyramidal}, and RANSAC homography estimation~\cite{fischler1981ransac}. At reference-supported grid points, let $a,b\in\mathbb{R}^2$ denote displacement under the reference and generated homographies. With $\epsilon=10^{-8}$ and $\tau=0.5$ pixels, define
\begin{equation}
  \begin{aligned}
    c(a,b) &= \max\!\left(0,
      \frac{a^\top b}{\max(\|a\|\|b\|,\epsilon)}\right), \\
    \rho(a,b) &= \min\!\left(
      \frac{\|a\|}{\max(\|b\|,\epsilon)},
      \frac{\|b\|}{\max(\|a\|,\epsilon)}\right).
  \end{aligned}
  \label{eq:cambgcomponents}
\end{equation}
The agreement at each grid point is
\begin{equation}
  A(a,b)=
  \begin{cases}
    c(a,b)\rho(a,b), & \|a\|\geq\tau, \\
    \exp[-(\|b\|/\tau)^2], & \|a\|<\tau.
  \end{cases}
  \label{eq:cambg}
\end{equation}
For moving reference regions, the score compares both displacement direction and magnitude. For nearly stationary regions, it treats reference motion as zero and penalizes generated displacement. A frame-pair score is the mean agreement over valid grid points multiplied by 100; a video score averages valid reference frame pairs. If reference motion is observable but generated motion estimation fails, the pair receives zero and remains in the denominator. Pairs with unobservable reference motion are excluded.

We detect at most 800 corners with quality threshold 0.01 and minimum spacing of 10 pixels. Lucas-Kanade tracking uses a $31\times31$ window, pyramid levels 0 to 3, and a forward-backward error of at most one pixel. A valid homography requires at least 40 consistent tracks, 24 inliers, an inlier ratio of 60\%, and support covering 8\% of the image and at least three quadrants. RANSAC uses a two-pixel threshold, confidence 0.995, at most 2,000 iterations. At least eight background points from a $12\times8$ grid must lie inside the inlier convex hull. Nonfinite transformations and displacements greater than 320 pixels are excluded. Cam-BG measures background motion consistency in the image plane.

{
  \small
  \bibliographystyle{unsrtnat}
  \bibliography{main}
}

\end{document}